\documentclass[runningheads]{llncs}

\usepackage[final,year=2024,ID=****]{eccv}

\usepackage{eccvabbrv}

\usepackage{graphicx}
\usepackage{booktabs}

\usepackage{epsfig}
\usepackage{graphicx}
\usepackage{amsmath}
\usepackage{amssymb}
\usepackage{multirow}
\usepackage{lipsum}
\usepackage{adjustbox}
\usepackage{booktabs}
\usepackage{makecell}
\usepackage{tabu}
\usepackage{bm}
\usepackage{xcolor}
\usepackage{comment}
\usepackage{colortbl}
\usepackage{color}
\usepackage{pifont}
\usepackage{colortbl}
\definecolor{Gray}{gray}{0.95}
\definecolor{Grayy}{gray}{0.6}

\newcommand{\paragrapha}[2][2pt]{\vspace{#1}\noindent\textbf{#2}}

\newcolumntype{C}[1]{>{\centering\arraybackslash}p{#1}}
\usepackage[accsupp]{axessibility}  

\usepackage[pagebackref,breaklinks,colorlinks,citecolor=eccvblue]{hyperref}

\usepackage{orcidlink}

\begin{document}

\title{Chain-of-Spot: Interactive Reasoning Improves \\ Large Vision-Language Models} 

\titlerunning{Chain-of-Spot}

\author{Zuyan Liu\inst{1}\thanks{Equal contribution. ~~\textsuperscript{\dag}Corresponding authors.} \and
Yuhao Dong\inst{1}$^{\star}$ \and
Yongming Rao\inst{2}$^{\dagger}$\and
Jie Zhou\inst{1}\and
Jiwen Lu\inst{1}$^{\dagger}$
}

\authorrunning{Liu et al.}

\institute{$^1$ Department of Automation, Tsinghua University ~~~ $^2$ Tencent}

\maketitle
\begin{figure}
  \centering
  \includegraphics[width=\linewidth]{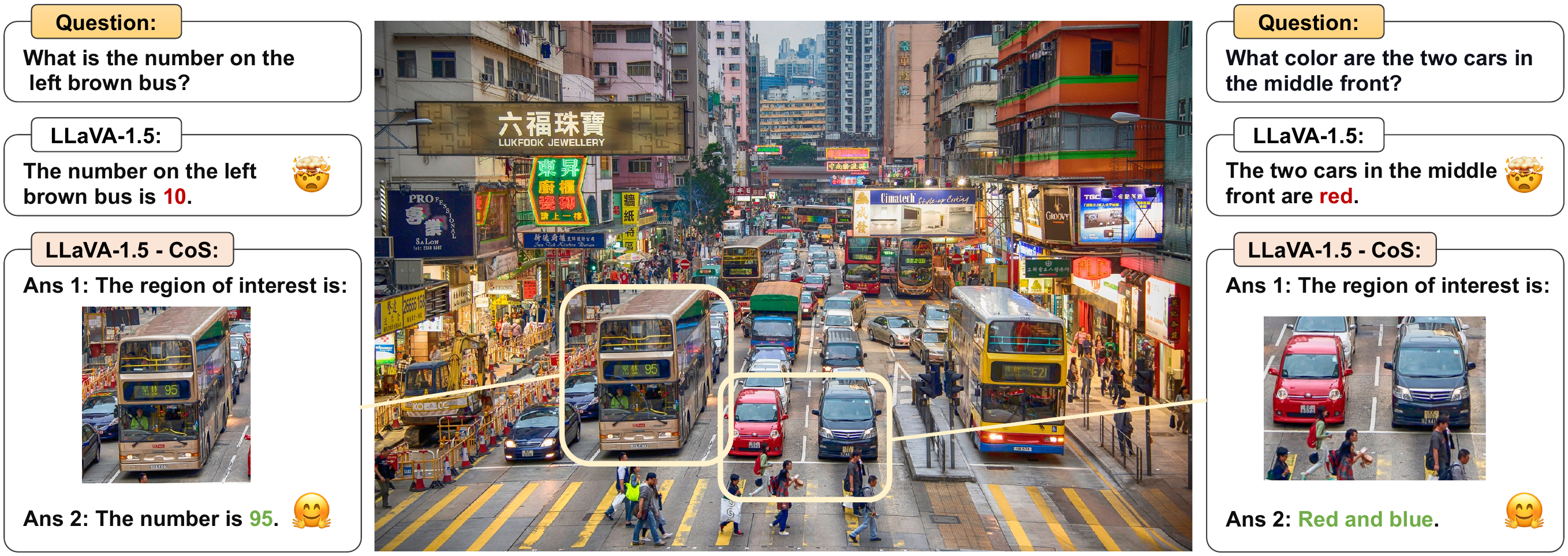}
  \caption{\textbf{Chain-of-Spot} encourages Large Vision-Language Models to identify the region of interest (ROI) in the image condition on the question and reasoning through an interactive manner, thereby improving the ability of visual understanding.}
  \label{fig:fig1}
\end{figure}

\begin{abstract}
In the realm of vision-language understanding, the proficiency of models in interpreting and reasoning over visual content has become a cornerstone for numerous applications. However, it is challenging for the visual encoder in Large Vision-Language Models (LVLMs) to extract useful features tailored to questions that aid the language model's response.  Furthermore, a common practice among existing LVLMs is to utilize lower-resolution images, which restricts the ability for visual recognition.  Our work introduces the \textbf{C}hain-\textbf{o}f-\textbf{S}pot (\textbf{CoS}) method, which we describe as Interactive Reasoning, a novel approach that enhances feature extraction by focusing on key regions of interest (ROI) within the image, corresponding to the posed questions or instructions.  This technique allows LVLMs to access more detailed visual information without altering the original image resolution, thereby offering multi-granularity image features.  By integrating Chain-of-Spot with instruct-following LLaVA-1.5 models, the process of image reasoning consistently improves performance across a wide range of multimodal datasets and benchmarks without bells and whistles and achieves new state-of-the-art results.  Our empirical findings demonstrate a significant improvement in LVLMs' ability to understand and reason about visual content, paving the way for more sophisticated visual instruction-following applications. Code and models are available at \url{https://github.com/dongyh20/Chain-of-Spot}. 
\end{abstract}

\section{Introduction}
\label{sec:intro}

The progress of large vision-language models (LVLMs)~\cite{llava, Bai2023QwenVLAF,li2023monkey} has attracted growing attention in the field of computer vision. Combining a visual encoder and a pre-trained large language model (LLM)~\cite{Touvron2023Llama2O, vicuna2023}, large vision-language models enable language models to extend their understanding beyond textual information to include visual data. This integration has opened up new possibilities where users can engage in natural language conversations about visual content, making the task increasingly important and popular across various domains. Despite the rapid development, substantial room exists to further improve models, including hallucinations, limited input resolution, and weak interpretability. 

Previous studies in large language models have shown that techniques such as the chain-of-thought~\cite{wei2022chain} represent a promising direction to enhance the reasoning capabilities of LLMs and improve the correctness of the generation results by guiding them through a structured thought process. Unlike LLMs, which primarily process and interpret text, LVLMs face the distinct challenge of comprehending and reasoning about the visual information presented in images. Therefore, it is natural to think: \textit{could LVLMs also be improved if we guide them to reason more deeply about the input visual content?}

A distinct and fundamental problem in visual understanding is the need to understand visual content at vastly different scales. Considering the diverse queries in visual instruction-following problems, it becomes even more challenging for the visual encoder of LVLMs to extract useful features containing various information that can help the language models understand and respond to various questions. Moreover, the computational burden imposed by processing an extensive number of tokens generated from high-resolution images has led to a common practice among existing LVLMs to utilize images of lower resolution, typically sized at 224x224 or 336x336 pixels. While recent methods like Monkey~\cite{li2023monkey} and LLaVA-HD~\cite{Liu2023ImprovedBW} have introduced multiple visual features to alleviate the problems, simply increasing the visual tokens will also bring considerable extra computation due to the long sequence length.

In this paper, we introduce an efficient, natural, and innovative approach to significantly enhancing the visual reasoning process of large vision-language models through Chain-of-Spot, which we term our method as Interactive Reasoning. We contend that the key to advancing LVLMs' ability to reason visually lies in instructing them to identify which parts of the given images are pertinent to answering a given question. After being equipped with the ability to identify the key region of interest (ROI) regarding the target questions or instructions, we zoom in on the critical part of the original image to offer LVLMs more detailed information. Therefore, we manage to perform multi-granularity image features while maintaining the image resolution invariant. To substantiate our hypothesis, we have redesigned the training and inference procedures of the widely adopted instruct-tuning pipelines. Specifically, we have restructured the question-answering framework within LVLMs in an interactive manner. Initially, we necessitate the LVLMs to pinpoint the region of interest relevant to the question and subsequently integrate both the global image features and localized region of interest (ROI) features to deduce the final answer. To instantiate this methodology, we have curated an enhanced version of the instruct-tuning dataset by annotating the region of interest (ROI) condition on the question by calculating the relevance map between language token attention and image features. Our findings indicate that this refined approach can considerably bolster the performance of vision-language models across a wide range of vision-language benchmarks.

We evaluate our idea of Chain-of-Spot by integrating our pipeline to improve the ability of the popular visual instruction tuning technique LLaVA~\cite{llava,Liu2023ImprovedBW}. We have observed a substantial leap in performance metrics after a single epoch of fine-tuning. Most notably, our enhanced version of LLaVA/13B achieves impressive results on a wide range of multimodal datasets. In the visual question answering tasks, accuracy is elevated from 80.0\% to 81.8\% with 1.8\% promotion on VQAv2~\cite{balanced_vqa_v2}. In GQA~\cite{hudson2019gqa}, we witness an increase from 63.3\% to 64.8\%, while in VizWiz~\cite{gurari2018vizwiz}, the accuracy rises markedly from 53.6\% to 58.0\%. The trend of improvement was consistent in multimodal benchmarks as well. For example, in SEEDBench~\cite{Li2023SEEDBenchBM}, the accuracy was augmented from 61.6\% to 62.3\%. Furthermore, when evaluating the long-context generation on the MM-Vet~\cite{Yu2023MMVetEL} dataset evaluated by the GPT-4~\cite{gpt4llm} model, we achieved an accuracy boost from 35.4\% to 37.6\%. A similar promotion is witnessed on the smaller vision-language instruct-following model LLaVA/7B, where we observe a similar lift on the multimodal benchmarks after fine-tuning with the idea of Chain-of-Spot. These results not only underscore the efficacy of Chain-of-Spot when applied to a state-of-the-art visual instruction tuning technique but also pave the way for future exploration of the reasoning mechanism of vision-language models.

\section{Related Work}

\paragrapha{Large Vision-Language Models. }Recent advancements in Large Language Models (LLMs)\cite{brown2020language,Ouyang2022TrainingLM,Touvron2023Llama2O,touvron2023llama}, including the notable ChatGPT\cite{OpenAI2023ChatGPT}, have showcased remarkable capabilities in language generation and interaction based on human prompts. The integration of LLMs with visual data has unlocked the potential for these models to not only comprehend image content but also recognize and interpret image components, thereby facilitating the emergence of Large Vision-Language Models (LVLMs). The BLIP series~\cite{Li2022BLIPBL,li2023blip,Dai2023InstructBLIPTG}, encompassing BLIP, BLIP2, and Instruction-BLIPs, has demonstrated that pre-trained language models are pivotal for visual-language tasks, addressing challenges in image captioning, visual question answering, and instruction-following. The KOSMOS series~\cite{peng2023kosmos,huang2023kosmos} and Shikra~\cite{chen2023shikra} have showcased LVLMs' proficiency in visual grounding and detection tasks through pre-training on interwoven image-text data. LLaVA~\cite{llava} recently introduced an approach to fine-tune open language models using GPT-generated visual instruction data, significantly enhancing practical instruction-following task performance. Subsequent works, including Qwen-VL~\cite{Bai2023QwenVLAF}, LLaVA-1.5~\cite{Liu2023ImprovedBW}, and mPLUG-Owl~\cite{ye2023mplug}, have continued to propel the capabilities of Large Vision-Language Models by implementing improved tuning strategies, expanding alignment data, and incorporating object-level annotations. Nevertheless, the efficacy of current LVLMs is hampered by the resolution of input images, as processing high-resolution imagery demands more robust image encoders and leads to a steep increase in computational costs. LLaVA-HD~\cite{Liu2023ImprovedBW} and Monkey~\cite{li2023monkey} have experimented with image patching techniques to handle higher resolutions, which is, however, costly and inefficient. Concurrent work~\cite{cao2024dualfocus} explores the gain from identifying macro and micro perspectives of images condition on questions while differing in fine-tuning data and inference operations. Our work seeks to introduce an efficient and effective approach through our novel concept of Chain-of-Spot, which guides the model to identify regions of interest within an image and interactively respond to queries, thereby addressing the challenges posed by high-resolution image processing.

\paragrapha{Visual Understanding with Multi-Scale Features. }The theme of visual understanding with multi-scale features focuses on effectively representing and processing visual information at multiple scales to improve the performance of various tasks. The Feature Pyramid Network (FPN)~\cite{lin2017feature} constructs semantic feature maps at multiple scales, enabling the detection of objects across a range of sizes with enhanced accuracy. This concept of a pyramidal hierarchy of features has been widely adopted and extended in various subsequent works~\cite{kirillov2019panoptic, cai2018cascade, zhu2020deformable}. The UNet architecture~\cite{ronneberger2015u} leverages a contracting path to assimilate contextual information by utilizing multi-scale features. High-Resolution Network (HRNet)~\cite{wang2020deep} maintains high-resolution representations throughout the network and effectively captures fine details by parallelly connecting convolutions across resolutions. Recent work~\cite{wang2020glance,wang2021adaptive,zhu2021dynamic,rao2023dynamic} find that utilizing multi-scale features can boost performance while improving recognition models' efficiency. Glance and Focus~\cite{wang2020glance} introduces a two-stage approach where the network first glances at the entire image to identify regions of interest and then focuses on these regions for detailed analysis, exploiting the benefits of multi-scale processing. However, regarding the meaningful usage of multi-scale features in the field of visual understanding, the method of exploiting multi-scale knowledge when generating vision-language response remains undiscovered. 

\section{Method}

In this section, we perform a detailed introduction to the idea of the Chain-of-Spot approach, an efficient, rational, and novel pipeline to effectively improve existing vision-language models' inference capabilities. We begin with a high-level overview of the motivation and idea of Chain-of-Spot, followed by a comprehensive description of the implementation pipeline of Chain-of-Spot, including data annotation, fine-tuning, and inference procedure on multimodal LLMs. 

\subsection{Overview of Chain-of-Spot }

\begin{figure}[tb]
  \centering
  \includegraphics[width=\linewidth]{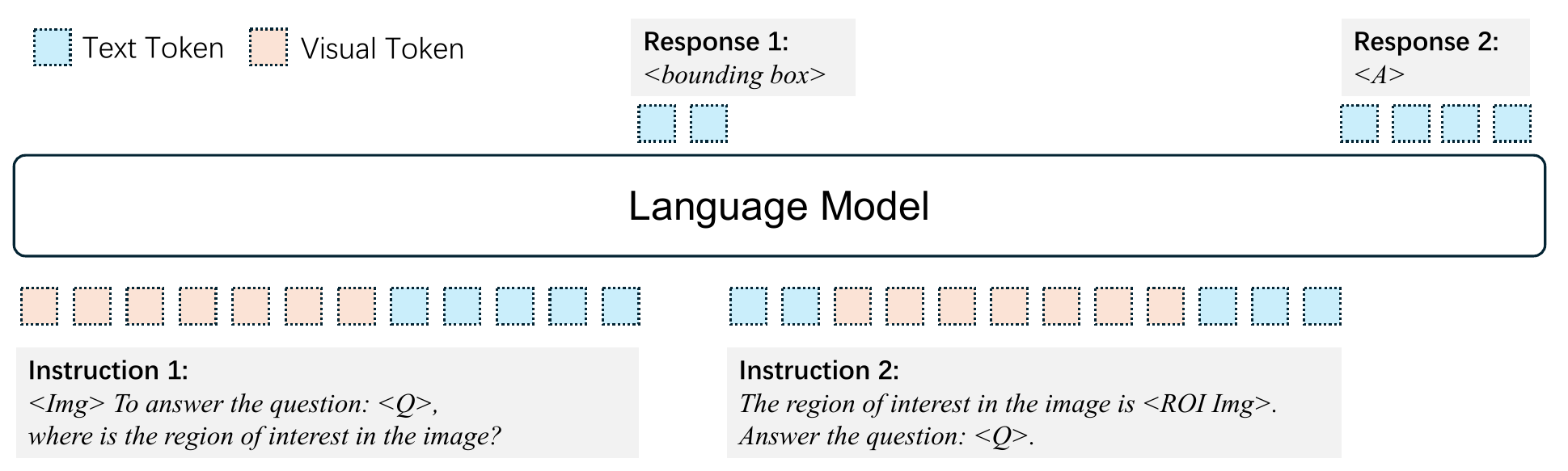}\vspace{-5pt}
  \caption{\textbf{Procedure of Chain-of-Spot. }We combine original visual tokens with questions in instruction 1 to ask LLMs to generate 
regions of interest (ROI) in the image. Then, we use the cropped image together with the questions again to generate better responses.  }
  \label{fig:intro}\vspace{-10pt}
\end{figure}

The integration of multimodal components within Large Vision-Language Models (LVLMs) typically involves a visual encoder, denoted as $\mathcal{E}_V$, and a language embedding encoder, represented by $\mathcal{E}_L$.  Initially, the model is presented with an image $I$ and an accompanying textual prompt $Q$, which could be a question or instruction.  The visual encoder $\mathcal{E}_V$, which is pre-trained, is employed to extract visual feature representations from $I$. Concurrently, the language component processes $Q$ by tokenizing and embedding the textual data into a vector space using the encoder $\mathcal{E}_L$.  The VLM then amalgamates these multimodal elements into a cohesive representation.  Subsequently, the model utilizes a transformer-based large language model $\mathcal{M}$, which has undergone fine-tuning across a spectrum of domains, to generate a pertinent textual response $R$.  The operational mechanism of the VLM, which underlies the response generation, can be mathematically expressed as follows:
\begin{equation}
    R=\mathcal{M}[\ \mathcal{E}_V(I),\ \mathcal{E}_L(Q)\ ]
\end{equation}
The impetus behind Chain-of-Spot is rooted in the processing methodology of Large Vision-Language Models (LVLMs). Conventional image encoders typically analyze an entire image, extracting a comprehensive global representation. This approach, however, often fails to effectively discern and prioritize the salient regions within the image that are most relevant to a given question.   We hypothesize that for a specific image $I$ and query $Q$, there exists an optimal region of interest $I_{C|Q}=\text{Crop}(I|Q)$ that holds the key to answering the question. By dynamically prompting the generation model to identify and concentrate on this critical region $I_{C|Q}$, we believe we can significantly enhance the model's reasoning capabilities. This interactive mechanism encourages a more focused and effective analysis, potentially leading to more accurate and contextually relevant responses.

Additionally, the length of image token embeddings is directly proportional to the square of the image resolution in contemporary LVLMs. Consequently, these models typically opt for lower image resolutions to mitigate excessive computational demands. We posit that this compromise on resolution may be inadequate for the performance of state-of-the-art LVLMs, potentially overlooking fine-grained details pivotal for addressing specific tasks. Strategies such as simply increasing the resolution of input images~\cite{Liu2023ImprovedBW} or employing structured high-resolution image patches~\cite{li2023monkey} to magnify and extract more nuanced information are not without drawbacks, primarily due to the resultant surge in computational complexity. Hence, preserving the original image resolution while selectively enlarging critical image regions could offer a more viable and efficient approach to enhancing detail without prohibitive computational costs.

Building upon the two aforementioned hypotheses, we have developed an interactive reasoning (Chain-of-Spot) approach that works in tandem with multi-scale images. This method dynamically concentrates on the prominent regions within the images $I_{C|Q}$, which are relevant to the posed questions $Q$. Such an approach is in harmony with the complex nature of human visual perception and cognitive processes. Starting with the original LVLMs with a baseline Large Language Model (LLM) denoted as $\mathcal{M}$, we fine-tune the VLM using a dataset comprising the centered-image $I_{C|Q}$, the question $Q$, and the original image $I$. Through this process, the model learns to identify the pivotal regions within the image. The result is an enhanced LLM, $\mathcal{M}_{\text{CoS}}$, which possesses interactive reasoning capabilities. During the inference phase, the interaction with $\mathcal{M}_{\text{CoS}}$ is initiated by instructing it to locate the key image area that corresponds to the question $Q$, as per a pre-defined instruction prompt, denoted as $\texttt{Inst.1}$. This step is formalized as follows:
\begin{equation}
    R(I_{C|Q})=\mathcal{M}_{\text{CoS}}[\ \mathcal{E}_V(I),\ \mathcal{E}_L(Q+\texttt{Inst.1})\ ]
    \label{eq:1}
\end{equation}
In response to $R(I_{C|Q})$, which includes the coordinates of the essential region, we crop the image to isolate $I_{C|Q}$. Subsequently, the response generation process utilizes both the original image $I$ and the cropped image $I_{C|Q}$, following the instruction prompt $\texttt{Inst.2}$:
\begin{equation}
    R=\mathcal{M}_{\text{CoS}}[\ \mathcal{E}_V(I, I_{C|Q}),\ \mathcal{E}_L(Q+\texttt{Inst.2})]
    \label{eq:2}
\end{equation}
It is important to note that the interactive generation unfolds conversationally, thereby exposing the entire reasoning process for identifying the key image region to the model. The procedure of Chain-of-Spot is illustrated in \cref{fig:intro}.

\subsection{Region of Interest Annotation via Relevance Propagation}

In this section, we introduce the pipeline we use to identify the region of interest in the image. We first show how we delimit the informative area in an image, and then introduce how we get the region of interest customized for a specific question based on the attention mechanism.

In order to detect potentially important areas in an image, we utilize an object detection model as our visual encoder to convert each image into several region features. Considering both speed and quality, Faster RCNN~\cite{ren2015faster} is selected as our detection model. Different from the traditional image encoder, the whole image is extracted into several bounding boxes containing the corresponding features of these regions. These features will then be forwarded into the language model $\mathcal{M}$ along with the embedded text tokens. 

Subsequently, we follow the rules proposed in ~\cite{chefer2021generic} to construct a relevance map $\Sigma$ from the attention map $\mathbf{A}$ which defines relations among all input tokens. The attention map is constructed based on the following rules:
\begin{equation}
    \mathbf{A}=\operatorname{softmax}\left(\frac{\mathbf{Q} \cdot \mathbf{K}^{\top}}{\sqrt{d_h}}\right) \\
    \label{eq:3}
\end{equation}
where $(\cdot)$ represents matrix multiplication, $\mathbf{Q}$ and $\mathbf{K}$ denote query matrix and key matrix, respectively. It should be recalled that if the attention mechanism is performed in a self-attention manner, the query and key are projected from input tokens, while in a cross-attention manner, the query comes from input text tokens, and the key comes from visual input. Based on the concept of residual connection, the relevance map $\Sigma$ is accumulated by adding up each layer's information contained in the attention map. For each layer, the process to unleash information obtained through the attention mechanism can be divided into two steps: first, we introduce attention interpreter $\Psi$ to reveal valuable clues contained in attention maps, which can be formatted as follows:
\begin{equation}
    \Psi=\mathbb{E}_h\left((\nabla \mathbf{A} \odot \mathbb{I}_{A>0}(\mathbf{A}))\right)
    \label{eq:4}
\end{equation}
where $\mathbb{E}_h$ denotes the average among multi-head attention, $\odot$ is Hadamard product, $\nabla$ denotes the back propagation of attention map $\mathbf{A}$ and $\mathbb{I}_{A>0}$ indicates only positive scores are considered. Then, we progressively update the relevance map following the rules:
\begin{equation}
    \Sigma = \Sigma + \Psi \cdot \Sigma
    \label{eq:5}
\end{equation}
After we obtain the relevance map, we post-process the relevance map to get the highlighted area in the image. Specifically, for each image region, we inquire and process the corresponding row in the relevance map to obtain a relevance score regarding the input image token. We then calculate a mask for all image regions and determine the highlighted area by a threshold $\epsilon$. We simply record the coordinates of this highlighted area as our training data.

\subsection{Training and Inference}

We introduce how to fine-tune the LLM for the capability of Chain-of-Spot (CoS) $\mathcal{M}_{\text{CoS}}$ from original LLM $\mathcal{M}$ with the prepared training data $(I, I_{C|Q}, Q)$: image $I$, the center of the image condition on question $I_{C|Q}$, and the question or instruction $Q$. 

As implemented in \cref{eq:1}, in the first step, we ask the language model $\mathcal{M}$ to find the region of interest condition on the question $Q$ and design the $\texttt{Inst.1}$:
\begin{equation}
\begin{aligned}
    \texttt{Inst.1: [}& I\texttt{] To answer the question: [}Q\texttt{],} \\
    &\texttt{where is the region of interest in the image?}
\end{aligned}
\end{equation}
To answer the \texttt{Inst.1}, we format the answer of the region in a bounding box manner based on the pre-processed image $I$ and $I_{C|Q}$, where we adopt four coordinates to represent the left, right boundary of the width $w_0, w_1$, the upper, bottom boundary of the height $h_0, h_1$, respectively. Note that we normalize the coordinates into $[0,1]$ scale and keep three decimal places for consistent training. We slightly extend the raw region of $I_{C|Q}$ to include more out-painting knowledge of the region of interest. The bounding box is treated directly as a string for the word embedding for the sake of simplicity. The answer to \texttt{Inst.1} can be formatted as:
\begin{equation}
    \texttt{Ans.1:str([}w_0,w_1,h_0,h_1\texttt{])}
\end{equation}
The second step is to answer the question after considering the region of interest. Given the part of input image condition on the question $I_{C|Q}$, We adopt the same pre-process procedure on $I_{C|Q}$ as image $I$ and treat the two images equally in large VLMs, which we zoom in the critical region of the image $I_{C|Q}$ to the same resolution as the original image $I$ and extract image representations using image encoder respectively. In \texttt{Inst.2}, we append the $I_{C|Q}$ and call back the raw question to emphasize the question to solve. The format of \texttt{Inst.2} is as follows:
\begin{equation}
\begin{aligned}
\texttt{Inst.2: }&\texttt{The region of interest in the image is [} I_{C|Q} \texttt{].} \\
&\texttt{Answer the question: [} Q \texttt{].}
\end{aligned}
\end{equation}
The answer to \texttt{Inst.2} is the original response. During the training phase, we concatenate \texttt{Inst.1}, \texttt{Inst.2} and the relevant answers for the forward process and adopt Cross-Entropy loss on answers. During the inference phase, we first feed-forward \texttt{Inst.1} into the vision-language model and let the model generate predictions on the bounding box of the region of interest. Then, we crop the image accordingly and formulate \texttt{Inst.2} for the vision-language model to adopt a normal response.

\section{Experiments}

In this section, we conduct extensive experiments to illustrate the improvement of the proposed Chain-of-Spot procedure on widely used multimodal vision-language models. By adopting Chain-of-Spot on state-of-the-art vision-language model LLaVA-1.5~\cite{Liu2023ImprovedBW}, we compare the improvements on a wide range of visual question answering and multimodal benchmarks and compare our models with existing large LVLMs. Subsequently, we perform an in-depth analysis of the design of the Chain-of-Spot. Additionally, we display quantitative results on the practical behavior of Chain-of-Spot and the comparisons with baselines. 

\subsection{Experimental Settings}

\paragrapha{Datasets and Baselines. }In our study, we employ a comprehensive suite of 11 multimodal datasets, each serving as a critical component in evaluating the performance of our proposed methods. These datasets are bifurcated into two distinct categories: visual question answering (VQA) and multimodal benchmarks. 

For visual question answering, we utilize six datasets: VQA-v2~\cite{balanced_vqa_v2}, GQA~\cite{hudson2019gqa}, VizWiz~\cite{gurari2018vizwiz}, Science QA~\cite{scienceqa}, TextVQA~\cite{textvqa}, and OKVQA~\cite{marino2019okvqa}. VQA-v2~\cite{balanced_vqa_v2} is a popular dataset that contains over 265K images from COCO~\cite{cocodataset} and abstract scenes with multiple questions. GQA~\cite{hudson2019gqa} offers a structured understanding of visual reasoning challenges with over 22M question-answer pairs grounded in 113,000 images. VizWiz~\cite{gurari2018vizwiz} is unique for its real-world images taken by visually impaired users, accompanied by nearly 31,000 question-answer pairs. Science QA~\cite{scienceqa} provides a specialized domain of scientific education with diagrams and associated questions to test knowledge-based VQA. TextVQA~\cite{textvqa} emphasizes the understanding of the text in images with about 28,000 images and 45,000 question-answer pairs. OKVQA~\cite{marino2019okvqa} extends the realm of VQA to general knowledge with 14K images that require external knowledge beyond the visual content. 

We integrate a suite of five diverse datasets to establish a comprehensive multimodal benchmark: SEED Bench~\cite{Li2023SEEDBenchBM}, MME~\cite{fu2023mme}, MMBench~\cite{liu2023mmbench}, POPE~\cite{Li2023EvaluatingOH}, and MM-Vet~\cite{Yu2023MMVetEL}. Specifically, SEED Bench~\cite{Li2023SEEDBenchBM} offers a rich multimodal analysis dataset consisting of synchronized multimodal sequences extracted from online video platforms. MME~\cite{fu2023mme} extends the benchmarking landscape with a broad array of 14 sub-tasks designed to evaluate multimodal learning comprehensively. MMBench~\cite{liu2023mmbench} focuses on assessing multimodal machine learning models, facilitating comparisons across a spectrum of tasks and data modalities. POPE~\cite{Li2023EvaluatingOH} presents a challenging dataset aimed at probing the hallucination phenomena in Language-Vision Language Models (LVLMs). Lastly, MM-Vet~\cite{Yu2023MMVetEL} is a platform for evaluating generative capabilities, with performance metrics benchmarked against the state-of-the-art GPT-4 model~\cite{OpenAI2023GPT4TR}.

To establish a strong benchmark for our experimental analysis, we adopt the state-of-the-art LLaVA-1.5~\cite{Liu2023ImprovedBW} method as our primary baseline. We adopt our methodology to two variants of the LLaVA architecture, LLaVA/7B and LLaVA/13B, and aim to demonstrate the scalability and generalizability of our approach across different model sizes.

\paragrapha{Implementation Details. }We delineate the specifics of our implementation. We augmented the supervised fine-tuning dataset from LLaVA-1.5~\cite{Liu2023ImprovedBW}, amassing a total of 665K images. It is important to note that multiple questions may be associated with a single image, culminating in approximately 3.3 million question-answer pairs. For the Chain-of-Spot fine-tuning phase, we select a random question-answer pair for each image to facilitate learning, ensuring the number of training steps per epoch remains consistent. Our starting point was the publicly accessible LLaVA-1.5/7B and LLaVA-1.5/13B models, which we fine-tuned using the instructions, image regions of interest (ROIs), and original images for one epoch. To maintain a baseline for comparison, we preserved the fine-tuning methodology of LLaVA-1.5 intact. The learning rate was set at 2e-5, with a global batch size of 1024. The maximum token length for the large language models was established at 2048 tokens. Our experimental setup included 8 NVIDIA A100 GPUs, utilizing the DeepSpeed Zero3~\cite{rasley2020deepspeed} optimization to minimize memory overhead. Regarding image preprocessing, we first padded the original images and their ROIs to a square format, followed by a resize to a resolution of $336 \times 336$ pixels. This preprocessing was necessary for feature extraction by the CLIP-ViT-L~\cite{radford2021learning} visual backbone. Throughout the fine-tuning process, we held the weights of the visual encoder and the multimodal projector constant while all language model parameters were trained in accordance with the established LLaVA-1.5 fine-tuning protocol.

\subsection{Main Results}

\begin{table}[t]
  \centering
  \caption{\textbf{Comparisons with vision-language models on visual question answering datasets. }Our Chain-of-Spot (CoS) consistently improves the vanilla LLaVA-1.5~\cite{Liu2023ImprovedBW} in all the benchmarks under different language model sizes. The best results are highlighted \textbf{bold} and the second are highlighted \underline{underline}.} \vspace{-5pt}
    \adjustbox{width=\linewidth}{
    \begin{tabular}{llcccccc}
    \toprule
    Method & Language~~~~ & ~~VQA$^{\text{V2}}$~~ & ~~GQA~~ & ~~VizWiz~~ & ~~SQA$^{\text{I}}$~~ & ~~VQA$^{\text{T}}$~~ & ~~OKVQA~~ \\
    \midrule
    BLIP-2~\cite{li2023blip} & Vicuna-13B & 65.0 & 32.3 & 19.6 & 61.0 & 42.5 & 45.9 \\
    InstructBLIP~\cite{Dai2023InstructBLIPTG} &Vicuna-13B&  - & 49.5 & 33.4 & 63.1 & 50.7 & - \\
    Shikra~\cite{chen2023shikra} & Vicuna-13B & 77.4 & - & - & - & - & 47.2 \\
    IDEFICS-80B~\cite{IDEFICS} & LLaMA-65B & 60.0 & 45.2 & 36.0 & - & 30.9 & - \\
    Qwen-VL~\cite{Bai2023QwenVLAF} & Qwen-7B & 79.5 & 59.3 & 35.2 & 67.1 & \textbf{63.8} & 58.6 \\
    Qwen-VL-Chat~\cite{Bai2023QwenVLAF} & Qwen-7B & 78.2 & 57.5 & 38.9 & 68.2 & 61.5 & 56.6 \\
    mPLUG-Owl2~\cite{ye2023mplug} & LLaMA-7B & 79.4 & 56.1 & 54.5 & 68.7 & 58.2 & 57.7 \\
    Monkey~\cite{li2023monkey} & Qwen-7B & 80.3 & 60.7 & \textbf{61.2} & 69.4 & - & \underline{61.3} \\
    \midrule
    LLaVA-1.5~\cite{Liu2023ImprovedBW}~~~ & Vicuna-7B~~~ & 78.5 & 62.0 & 50.0 & 66.8 & 58.2 & 57.9 \\
    \rowcolor{Gray} LLaVA-1.5+\textbf{CoS}~~~ & Vicuna-7B~~~ & \underline{80.7} & \underline{63.7} & 50.8 & 68.2 & 60.9 & 58.4 \\
    \midrule
    LLaVA-1.5~\cite{Liu2023ImprovedBW}~~~ & Vicuna-13B~~~ & 80.0  & 63.3 & 53.6 & \underline{71.6} & 61.3  & 60.9 \\
    \rowcolor{Gray} LLaVA-1.5+\textbf{CoS}~~~ & Vicuna-13B~~~ & \textbf{81.8} & \textbf{64.8} & \underline{58.0} & \textbf{71.9} & \underline{62.4} & \textbf{62.9} \\
    \bottomrule
    \end{tabular}%
    }\vspace{-10pt}
  \label{tab:sota1}%
\end{table}%

We perform our main experiments on 11 widely used and challenging multimodal benchmarks. We clearly show the performance compared with our baseline LLaVA-1.5 and the comparisons with other vision-language models to show the superiority of our method. 

\paragrapha{Results on Visual Question Answering Datasets. }We rigorously evaluate the effectiveness of our Chain-of-Spot approach through extensive experiments on six challenging datasets that are widely recognized in the visual question-answering research community: VQA-v2~\cite{balanced_vqa_v2}, GQA~\cite{hudson2019gqa}, VizWiz~\cite{gurari2018vizwiz}, ScienceQA~\cite{scienceqa}, TextVQA~\cite{textvqa}, and OKVQA~\cite{marino2019okvqa}. Results are shown in \cref{tab:sota1}. Upon integrating our Chain-of-Spot method with the LLaVA/13B model, we observed a remarkable enhancement in performance, surpassing all baseline metrics on the aforementioned datasets, setting new state-of-the-art results on four of them.  This improvement was most pronounced in the VizWiz dataset, where we achieved a 4.4\% increase.  On the more general VQA-v2 and GQA datasets, we saw increases of 1.8\% and 1.5\%, respectively. The performance on ScienceQA, TextVQA, and OKVQA datasets, with improvements of 0.3\%, 1.1\%, and 2.0\%, respectively, further demonstrates the versatility of our approach. Similarly, the LLaVA/7B model, enhanced with Chain-of-Spot, consistently outperformed the baseline across all six benchmarks with gains of 2.2\% on VQA-v2, 1.7\% on GQA, 0.8\% on VizWiz, 1.4\% on ScienceQA, 2.7\% on TextVQA, and 0.5\% on OKVQA. Notably, the fine-tuned LLaVA/7B model using our Chain-of-Spot approach achieved superior performance to the baselines of the LLaVA/13B model on the VQA-v2 and GQA datasets, which exemplifies the potential of our method to achieve state-of-the-art performance even when computational resources are limited. This robust performance underscores the efficacy of our proposed method and highlights its potential to enhance visual question-answering capabilities significantly.

\paragrapha{Results on Multimodal Benchmarks. }We evaluated our innovative Chain-of-Spot method across five multimodal benchmarks specifically designed to test the limits of multimodal understanding and reasoning. The benchmarks included SEEDBench~\cite{Li2023SEEDBenchBM}, MME~\cite{fu2023mme}, MMBench~\cite{liu2023mmbench}, POPE~\cite{Li2023EvaluatingOH}, and MM-Vet~\cite{Yu2023MMVetEL}, each presenting its own challenges and requiring a nuanced understanding of multimodal inputs. We reported the overall accuracy and dissected the performance on image-only data to capture the full scope of our method's capabilities on SEEDBench. Results are shown in \cref{tab:sota2}. When implemented on the LLaVA-1.5/13B models, our Chain-of-Spot method consistently outperformed the established baselines across all six benchmarks, demonstrating the robustness and versatility of our approach.   On SEEDBench, we observed a solid gain of 0.7\%, while the image-only subset of SEEDBench saw a more significant improvement of 1.4\%. The MMBench and MM-Vet benchmarks also showed notable improvements of 0.5\% and 2.2\%, respectively. When we adapted Chain-of-Spot to the smaller LLaVA-1.5 model with a 7B language model as its backbone, we still observed similar gains across the five benchmarks, which is a testament to our method's scalability and effectiveness across different model sizes. Most impressively, with Chain-of-Spot, LLaVA-1.5 models achieved state-of-the-art results on all benchmarks, firmly establishing our proposed method as a significant step forward in multimodal learning.

\begin{table}[t]
  \centering
  \caption{\textbf{Comparisons with vision-language models on multimodal benchmarks. }LLaVA-1.5~\cite{Liu2023ImprovedBW} with Chain-of-Spot (CoS) achieves state-of-the-art performance on all the multimodal benchmarks, surpassing previous LVLMs by a large margin. The best results are highlighted \textbf{bold} and the second are highlighted \underline{underline}.} \vspace{-5pt}
    \adjustbox{width=\linewidth}{
    \begin{tabular}{llcccccc}
    \toprule
    Method & Language~~~~ & ~~SEED~~ & ~~SEED$^{I}$~~ & ~~MME~~ & ~~MMB~~ & ~~POPE~~ & ~~MM-Vet~~ \\
    \midrule
    BLIP-2~\cite{li2023blip} & Vicuna-13B & 46.4 & - & 1293.8 & - & 85.3 & 22.4 \\
    InstructBLIP~\cite{Dai2023InstructBLIPTG} & Vicuna-13B & 53.4 & 58.8 & 1212.8 & 36.0 & 78.9 & 25.6 \\
    Qwen-VL~\cite{Bai2023QwenVLAF} & Qwen-7B & 56.3 & 62.3 & - & 38.2 & - & - \\
    Qwen-VL-Chat~\cite{Bai2023QwenVLAF} & Qwen-7B & 58.2 & 65.4 & 1487.5 & 60.6 & - & -\\
    mPLUG-Owl2~\cite{ye2023mplug} & LLaMA-7B & \underline{61.6} & - & 1450.2 & 64.5 & - & \underline{36.2} \\
    \midrule
    LLaVA-1.5~\cite{Liu2023ImprovedBW}~~~ & Vicuna-7B~~~ & 58.6 & 66.1 & 1510.7 & 64.3 & 85.9 & 30.5 \\
    \rowcolor{Gray}  LLaVA-1.5+\textbf{CoS}~~~ & Vicuna-7B~~~ & 59.7 & 67.1 & 1501.1 & 64.4 & \textbf{86.4} & 30.8\\
    \midrule
    LLaVA-1.5~\cite{Liu2023ImprovedBW}~~~ & Vicuna-13B~~~ & \underline{61.6}  & \underline{68.2} & \underline{1531.3}  & \underline{67.7}  & 85.9 & 35.4\\
    \rowcolor{Gray} LLaVA-1.5+\textbf{CoS}~~~ & Vicuna-13B~~~ & \textbf{62.3} & \textbf{69.6} & \textbf{1546.1} & \textbf{68.2} & \underline{86.1} & \textbf{37.6}\\
    \bottomrule
    \end{tabular}%
    }\vspace{-10pt}
  \label{tab:sota2}%
\end{table}%

\subsection{Analysis}

In the analysis part, we conduct convincing experiments to demonstrate the key design of the idea of Chain-of-Spot. Additionally, we perform abundant visualizations to serve as a supplement to our impressive quantitative improvement. 

\begin{table}[t]
\caption{\textbf{Analysis and Ablations.} We conduct ablations on the choices of reasoning techniques and training strategies to demonstrate the effectiveness of Chain-of-Spot as well as provide more insights into our method. w/o Chain-of-Spot indicates not implementing chain-of-spot method while only continuously fine-tuning LLaVA-1.5 for one epoch. } \vspace{-15pt}
\begin{subtable}{.5\linewidth}
\caption{Choices of Reasoning Techniques}
\label{tab:ablation_1}
\adjustbox{width=\linewidth}{
\begin{tabular}{lcc}
\toprule
Reasoning Techniques~~~ & ~~~GQA~~~ & ~~~VQA$^{V2}$~~~ \\
\midrule
w/o Chain-of-Spot  & 63.9  & 80.2 \\
\midrule
Center Crop  & 64.0  & 80.4 \\
From Image & 64.1 & 80.6 \\
From Questions & 64.3 & 81.1\\
\rowcolor{Gray} From Single Question & \textbf{64.8} & \textbf{81.8} \\
\bottomrule
\end{tabular}
}
\end{subtable}
\begin{subtable}{.5\linewidth}
\caption{Ablations on Training Strategies}
\label{tab:ablation_2}
\adjustbox{width=\linewidth}{
\begin{tabular}{lcc}
\toprule
Training Strategies~~~~~~~ & ~~~GQA~~~ & ~~~VQA$^{V2}$~~~ \\
\midrule
\rowcolor{Gray}$\text{Learning Rate}=1.0\times$  & \textbf{64.8}  & \textbf{81.8} \\
$\text{Learning Rate}=0.1\times$ & 64.1  & 81.2 \\
\midrule
\rowcolor{Gray}$\text{Epochs}=1$ & 64.8 & \textbf{81.8}\\
$\text{Epochs}=2$ & \textbf{64.9} & 81.2 \\
$\text{Epochs}=3$ & 64.4 & 80.9\\
\bottomrule
\end{tabular}
}
\end{subtable}
\vspace{-15pt}
\end{table}

\paragrapha{Choices of Reasoning Techniques. }In the method section of our paper, we introduce a novel approach to generating regions of interest (ROI) in images by leveraging one-to-one correspondence between question-answer pairs.  We conducted a series of analytical experiments in \cref{tab:ablation_1} to validate the critical importance of this design choice.  Initially, we trained our LLaVA model continuously for one epoch without the Chain-of-Spot component to establish a baseline.  Subsequently, we explored various techniques for ROI determination, including (1) a naive method that crops and zooms into the image center, (2) an uninformed prediction model that identifies ROIs directly from the image without language cues, (3) a broad approach using all questions in the LLaVA dataset to elicit a generalized response, and (4) a targeted prediction based on individual questions. The comparative results clearly demonstrate that our method outperforms these alternatives, underscoring the efficacy of our question-answer pair-driven ROI generation technique in achieving superior performance. Note that all the alternatives of Chain-of-Spot outperform the baselines that further train one epoch without Chain-of-Spot, which shows the idea of instructing the model to reason works effectively. 

\paragrapha{Ablations on Training Strategies. }In the process of fine-tuning the LLaVA model, we conducted an exploration of hyper-parameter configurations in \cref{tab:ablation_2}. Notably, a key parameter adjustment involved experimenting with a reduced learning rate.  However, experimental evidence suggested that retaining the original learning rate was more beneficial, potentially attributable to the necessity for stronger learning gradients during the critical task of identifying regions of interest (ROI) within the learning procedure.    Furthermore, the number of training epochs was scrutinized, revealing that prolonging the training duration did not correspond to an enhancement in model performance.    This plateau in performance gains beyond a certain number of epochs is likely a consequence of the LLaVA dataset's limited size, which introduces a heightened risk of overfitting when subjected to excessive training.

\begin{figure}[tb]
  \centering
  \includegraphics[width=\linewidth]{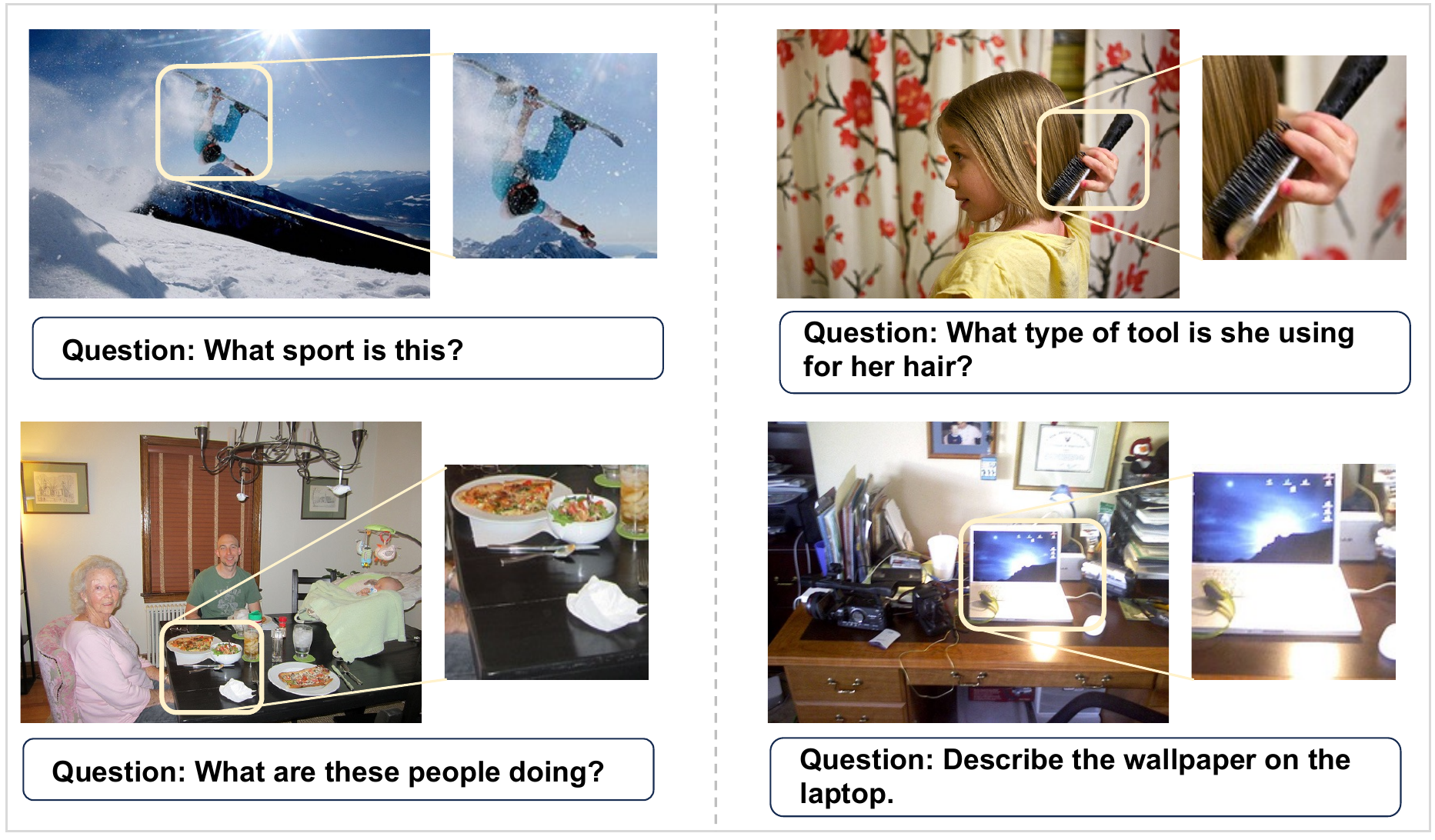}\vspace{-5pt}
  \caption{\textbf{Visualizations on Chain-of-Spot. }Chain-of-Spot shows the reasonable region of interest condition on the given questions. }
  \label{fig:viz1}\vspace{-15pt}
\end{figure}

\paragrapha{Visualizations on Chain-of-Spot. }We employ visualizations in \cref{fig:viz1} to elucidate the effectiveness of the Chain-of-Spot approach in identifying regions of interest pertinent to query responses within images. For instance, in the visualization of the top-left image, when prompted with the question 'What sport is this?', the model adeptly narrows in on the individuals engaged in skiing, thereby isolating the key area that confirms the sport in question. Similarly, in the bottom-left image, under the question 'What are these people doing?', the model does not directly identify the persons in the image while hones in on a dish situated on a table, which in turn unambiguously addresses the posed question. Moving to the top-right image, the model's focus shifts to a comb held by a girl, which stands as the definitive answer to the associated question. In the case of the bottom-right image, the model demonstrates its perceptual acuity by recognizing the wallpaper pattern and, precisely, identifies a laptop as the answer to the question at hand in a messy room with complex disturbance. These visualizations collectively showcase the model's remarkable capacity to selectively crop and spotlight the most salient region within an image that directly pertains to the answer, thereby affirming its robust capability for reasoning and understanding within visual contexts.

\begin{figure}[tb]
  \centering
  \includegraphics[width=\linewidth]{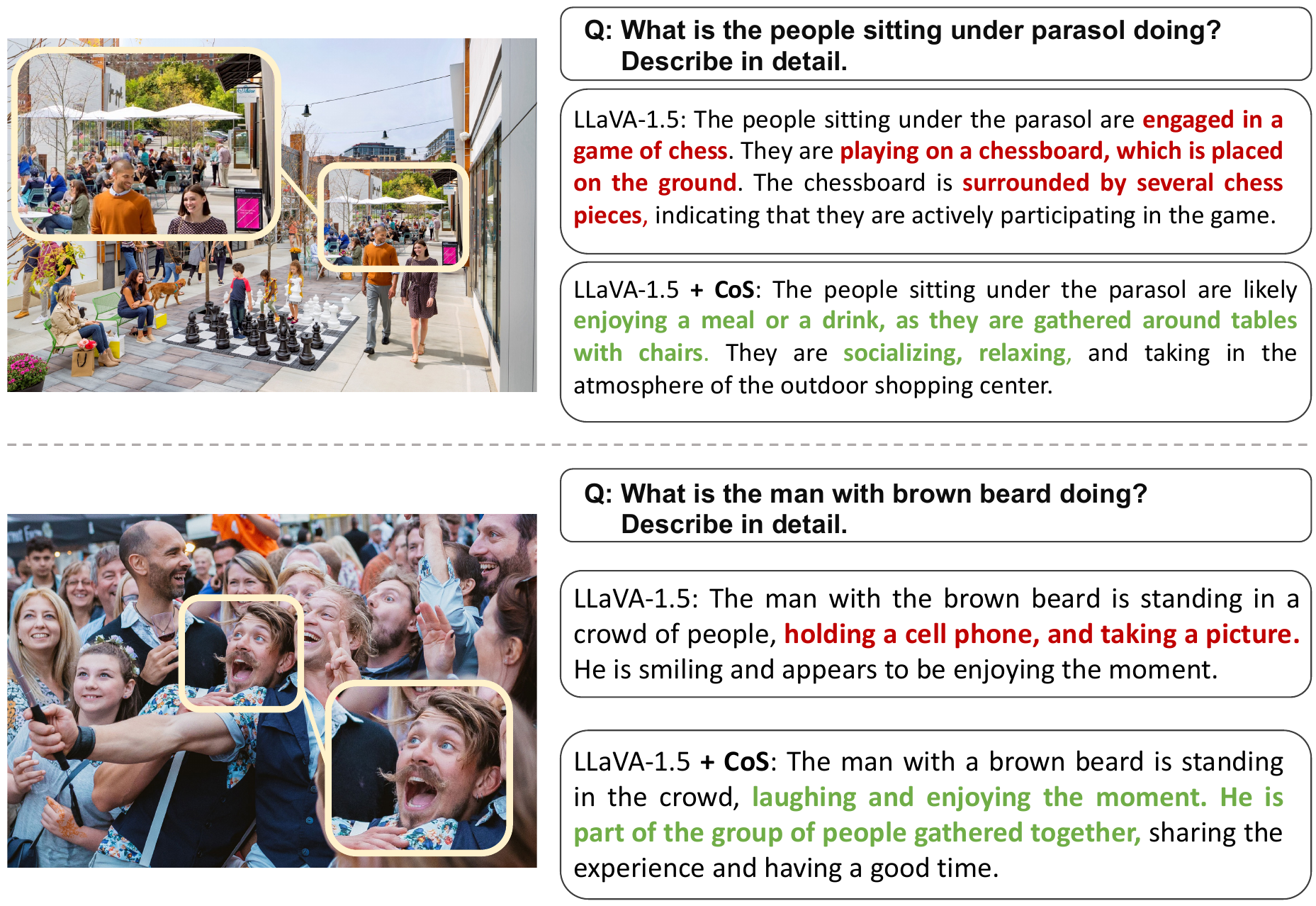}\vspace{-5pt}
  \caption{\textbf{Generation comparisons after implementing Chain-of-Spot. }Chain-of-Spot corrects the focus and the answers of the LLaVA model on complex visual question cases. }
  \label{fig:viz2}\vspace{-15pt}
\end{figure}

\paragrapha{Qualitative Comparisons of Chain-of-Spot. }We present qualitative comparisons in \cref{fig:viz2} to demonstrate the enhancements brought about by implementing Chain-of-Spot within the LLaVA-1.5 framework.  Specifically, we detail two illustrative examples that underscore the advancement of the reasoning process.  For the initial example, we tasked the model with identifying a person located in a distant position (beneath a parasol).  Prior to the integration of Chain-of-Spot, LLaVA-1.5 incorrectly focused on the nearby children engaged in a game of chess.  However, with our Chain-of-Spot approach, the model is able to discern the significance of the parasol, thereby accurately zooming in on the more remote area and providing the correct response that the individuals are involved in activities such as eating, drinking, and conversing.  The second example showcases a complex scenario featuring many individuals, where we highlight a specific person distinguished by a brown beard.  Without the aid of Chain-of-Spot, LLaVA-1.5 confounds this individual with someone in close proximity and inaccurately reports that he is holding a cell phone and taking a photo.  In contrast, by employing Chain-of-Spot, the model is adept at isolating the correct man and accurately depicting his actions.  These examples collectively illustrate the significant strides our approach makes in refining the reasoning capabilities of LLaVA-1.5, thereby enhancing its performance in complex visual recognition tasks.

\section{Conclusion, Limitations and Societal Impact}

In this paper, we have presented an innovative approach Chain-of-Spot to improve large visual-language models' reasoning and understanding ability effectively. By equipping LVLMs with the ability to identify and focus on these key regions of interest (ROI), we provide a pathway to more detailed visual information without compromising image resolution. We have reimagined the training and inference procedures within the established instruct-tuning pipelines, introducing an interactive question-answering framework that compels LVLMs to integrate global image and localized ROI features. Our empirical results underscore the effectiveness of our approach, demonstrating marked improvements in LVLM performance across various vision-language benchmarks.

One limitation of our approach may come from the insufficient amount of training data for fine-tuning, which may cause inadequate guidance for the LVLMs to find the ROI in the image. The enhancement of LVLMs can profoundly impact society by improving assistive technologies and smart automation, yet it also necessitates careful consideration of ethical implications related to privacy and the increased potential for sophisticated surveillance systems.

\begin{appendix}

\section{More Visualizations}

In this supplementary section, we follow the visualization framework outlined in the main manuscript to present a broader array of qualitative results under a variety of scenarios, thereby providing a more comprehensive understanding of our model's performance. By conducting an in-depth analysis of the regions of interest (ROIs) across diverse situations, we demonstrate the reasoning capabilities of our approach. Additionally, we perform more comparisons against our LLaVA~\cite{Liu2023ImprovedBW} baselines under visual question-answering datasets. 

\begin{figure}
  \centering
  \includegraphics[width=\linewidth]{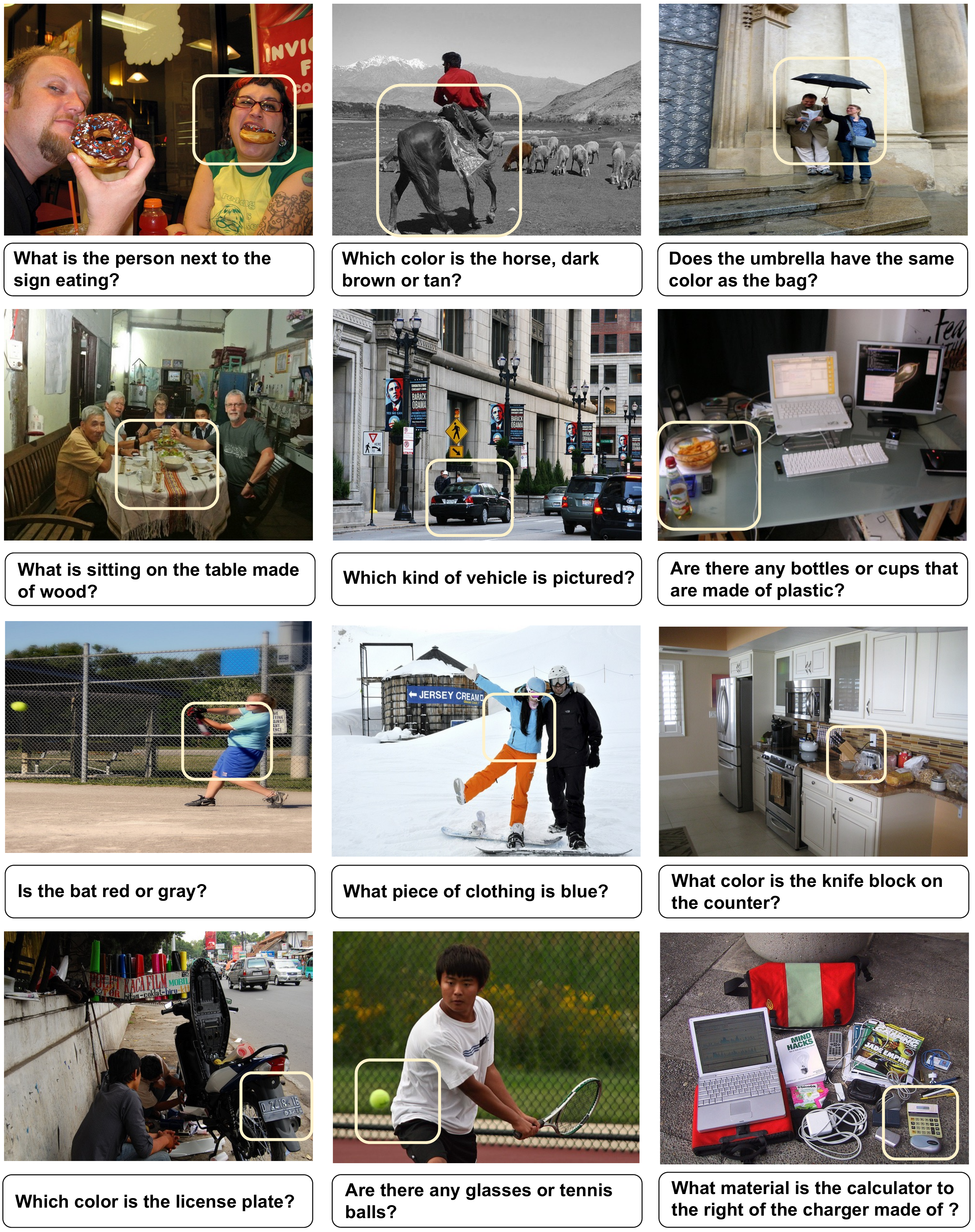}
  \caption{\textbf{More Results of Chain-of-Spot. }We illustrate the question and the relevant region of interest in the image marked by Chain-of-Spot. }
  \label{fig:supp2}
\end{figure}

\begin{figure}
  \centering
  \includegraphics[width=\linewidth]{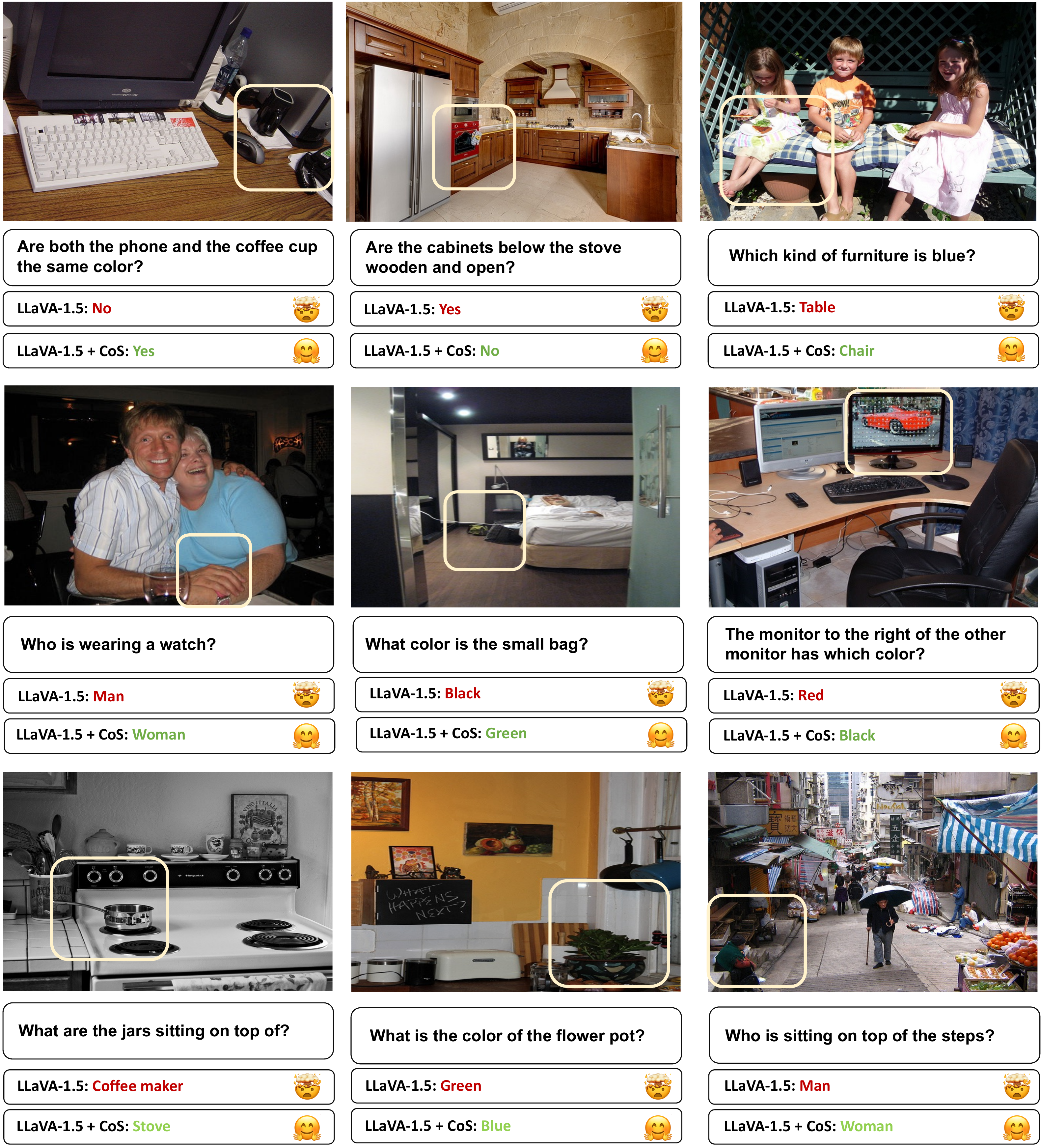}
  \caption{\textbf{More Comparisons with Baselines. }Results before and after Chain-of-Spot are illustrated as LLaVA-1.5 and LLaVA-1.5+CoS, respectively. The correct answers are colored green and the wrong answers are colored red. We mark out the region of interest identified by Chain-of-Spot.}
  \label{fig:supp1}
\end{figure}

\subsection{More Results of Chain-of-Spot}

We perform abundant visualization results from GQA~\cite{hudson2019gqa} datasets to clearly illustrate the relationship between the question or instruction and the region of interest identified by Chain-of-Spot. Results are shown in \cref{fig:supp2}. Chain-of-Spot can identify the critical object (the calculator to the right of the charger) among complex objects in the subfigure (row 4th, column 3rd), concentrate on the small objects (the license plate) in the whole image in the subfigure (row 4th, column 1st), point out the correct target (the person next to the sign) under similar disturbance in the subfigure (row 1st, column 1st).

\subsection{More Comparisons with Baselines} 

In this supplementary section of our study, we extend our analysis through a comprehensive set of comparisons against baselines, where we employ our comparisons on the GQA dataset~\cite{hudson2019gqa}. Results are illustrated in \cref{fig:supp1}. We show the generated response both before and after employing our Chain-of-Spot fine-tuning. By visually annotating the regions of interest (ROI) identified by Chain-of-Spot on the images, we provide a clear and intuitive understanding of how the model's focus is refined.  

Notably, we observe a marked enhancement in the performance of the underlying Large Vision-language models post Chain-of-Spot intervention. This improvement is attributed to the module's ability to excise extraneous visual information and concentrate on the salient image regions pertinent to the posed question. The resultant cropping of the ROI effectively mitigates the distraction caused by irrelevant image content, thereby sharpening the model's attention and significantly elevating the accuracy of the responses. For example, in the subfigure (row 2nd, column 3rd), we can observe that LLaVA baseline mistakes the wallpaper in the monitor as the color of the monitor, however, zooming in to the monitor can effectively identify the correct color of the monitor. In the subfigure (row 2nd, column 1st), the hands of the two persons in the image have complex relationships, resulting in mistakes in the LLaVA baseline. Chain-of-Spot can focus on the hand region in the image, which offers more clear information to the language model. 

\section{Statistical Analysis of Chain-of-Spot}

\begin{figure}[t]
  \centering
  \includegraphics[width=0.5\linewidth]{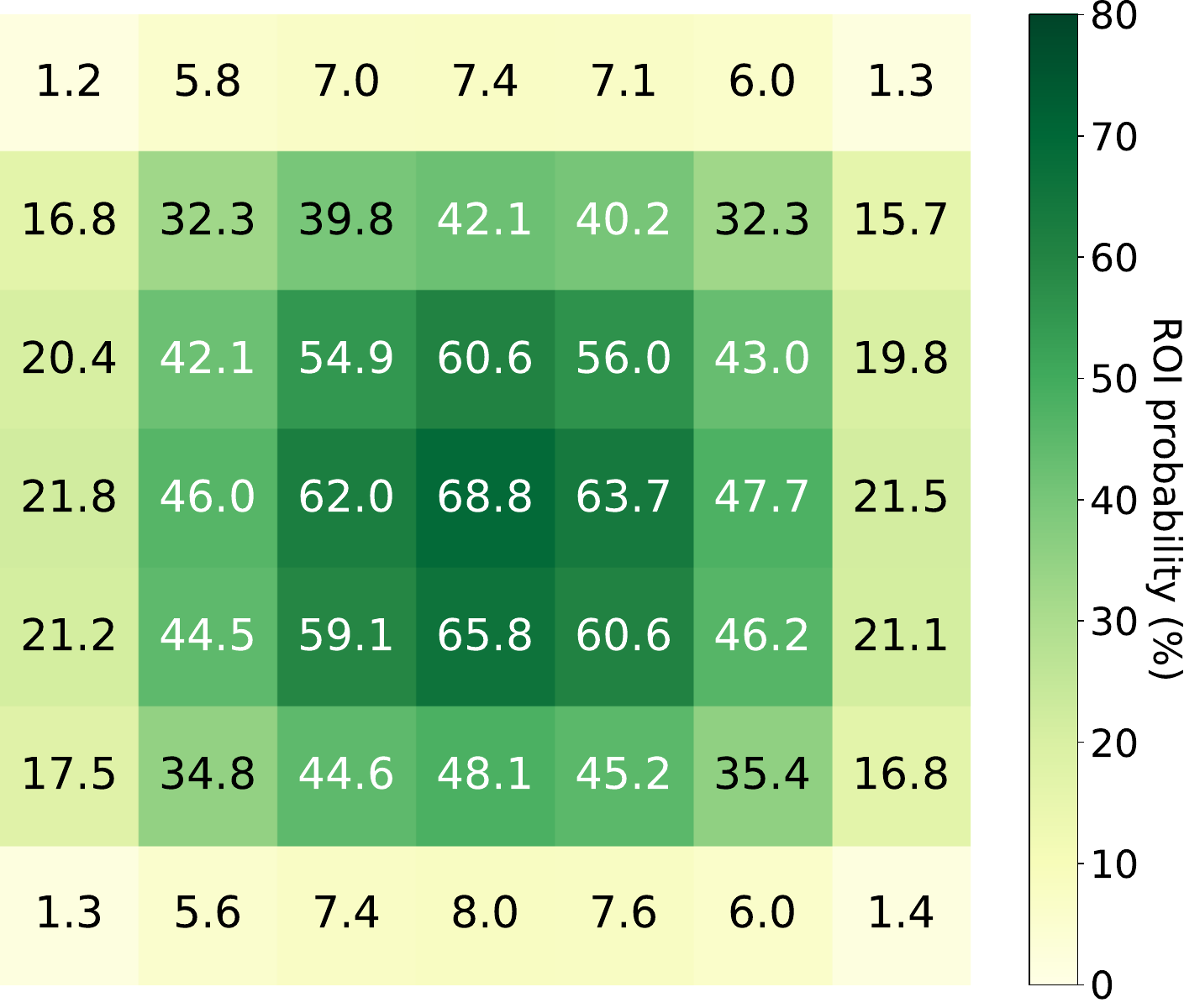}
  \caption{\textbf{Statistical Analysis of Chain-of-Spot. }We show the statistical results of the ROI in all the question-answer pairs of LLaVA~\cite{Liu2023ImprovedBW} dataset. We can observe that the center of the image contributes most to the information. }
  \label{fig:stat}
\end{figure}

We are interested in examining the statistical outcomes associated with the distribution of the region of interest (ROI) within various datasets.  To this end, we present an analysis of the reasoning outcomes derived from all question-answering pairs under consideration. Results are shown in \cref{fig:stat}. Our findings indicate that the image center plays a predominant role in influencing performance, with the critical region encompassing roughly one-quarter of the entire image area.

\end{appendix}

%
%
\bibliographystyle{splncs04}
\bibliography{main}

\begin{thebibliography}{10}
\providecommand{\url}[1]{\texttt{#1}}
\providecommand{\urlprefix}{URL }
\providecommand{\doi}[1]{https://doi.org/#1}

\bibitem{Bai2023QwenVLAF}
Bai, J., Bai, S., Yang, S., Wang, S., Tan, S., Wang, P., Lin, J., Zhou, C., Zhou, J.: Qwen-vl: A frontier large vision-language model with versatile abilities. ArXiv:2308.12966  (2023)

\bibitem{brown2020language}
Brown, T., Mann, B., Ryder, N., Subbiah, M., Kaplan, J.D., Dhariwal, P., Neelakantan, A., Shyam, P., Sastry, G., Askell, A., et~al.: Language models are few-shot learners. Advances in neural information processing systems pp. 33, 1877--1901 (2020)

\bibitem{cai2018cascade}
Cai, Z., Vasconcelos, N.: Cascade r-cnn: Delving into high quality object detection. In: Proceedings of the IEEE conference on computer vision and pattern recognition. pp. 6154--6162 (2018)

\bibitem{cao2024dualfocus}
Cao, Y., Zhang, P., Dong, X., Lin, D., Wang, J.: Dualfocus: Integrating macro and micro perspectives in multi-modal large language models. arXiv preprint arXiv:2402.14767  (2024)

\bibitem{chefer2021generic}
Chefer, H., Gur, S., Wolf, L.: Generic attention-model explainability for interpreting bi-modal and encoder-decoder transformers. In: Proceedings of the IEEE/CVF International Conference on Computer Vision. pp. 397--406 (2021)

\bibitem{chen2023shikra}
Chen, K., Zhang, Z., Zeng, W., Zhang, R., Zhu, F., Zhao, R.: Shikra: Unleashing multimodal llm's referential dialogue magic. arXiv preprint arXiv:2306.15195  (2023)

\bibitem{vicuna2023}
Chiang, W.L., Li, Z., Lin, Z., Sheng, Y., Wu, Z., Zhang, H., Zheng, L., Zhuang, S., Zhuang, Y., Gonzalez, J.E., Stoica, I., Xing, E.P.: Vicuna: An open-source chatbot impressing gpt-4 with 90\%* chatgpt quality. \url{https://lmsys.org/blog/2023-03-30-vicuna/} (March 2023)

\bibitem{Dai2023InstructBLIPTG}
Dai, W., Li, J., Li, D., Tiong, A.M.H., Zhao, J., Wang, W., Li, B.A., Fung, P., Hoi, S.C.H.: Instructblip: Towards general-purpose vision-language models with instruction tuning. ArXiv:2305.06500  (2023)

\bibitem{fu2023mme}
Fu, C., Chen, P., Shen, Y., Qin, Y., Zhang, M., Lin, X., Qiu, Z., Lin, W., Yang, J., Zheng, X., et~al.: Mme: A comprehensive evaluation benchmark for multimodal large language models. arXiv preprint arXiv:2306.13394  (2023)

\bibitem{balanced_vqa_v2}
Goyal, Y., Khot, T., Summers{-}Stay, D., Batra, D., Parikh, D.: Making the {V} in {VQA} matter: Elevating the role of image understanding in {V}isual {Q}uestion {A}nswering. In: Conference on Computer Vision and Pattern Recognition (CVPR) (2017)

\bibitem{gurari2018vizwiz}
Gurari, D., Li, Q., Stangl, A.J., Guo, A., Lin, C., Grauman, K., Luo, J., Bigham, J.P.: Vizwiz grand challenge: Answering visual questions from blind people. In: Proceedings of the IEEE conference on computer vision and pattern recognition. pp. 3608--3617 (2018)

\bibitem{huang2023kosmos}
Huang, S., Dong, L., Wang, W., Hao, Y., Singhal, S., Ma, S., Lv, T., Cui, L., Mohammed, O.K., Liu, Q., et~al.: Language is not all you need: Aligning perception with language models. arXiv preprint arXiv:2302.14045  (2023)

\bibitem{hudson2019gqa}
Hudson, D.A., Manning, C.D.: Gqa: A new dataset for real-world visual reasoning and compositional question answering. In: Proceedings of the IEEE/CVF conference on computer vision and pattern recognition. pp. 6700--6709 (2019)

\bibitem{IDEFICS}
HugoLaurencon, van Strien, D., Bekman, S., Tronchoon, L., Saulnier, L.: Introducing idefics: An open reproduction of state-of-the-art visual language model. \url{https://huggingface.co/blog/idefics} (2023)

\bibitem{kirillov2019panoptic}
Kirillov, A., Girshick, R., He, K., Doll{\'a}r, P.: Panoptic feature pyramid networks. In: Proceedings of the IEEE/CVF conference on computer vision and pattern recognition. pp. 6399--6408 (2019)

\bibitem{Li2023SEEDBenchBM}
Li, B., Wang, R., Wang, G., Ge, Y., Ge, Y., Shan, Y.: Seed-bench: Benchmarking multimodal llms with generative comprehension. ArXiv:abs/2307.16125  (2023)

\bibitem{li2023blip}
Li, J., Li, D., Savarese, S., Hoi, S.: Blip-2: Bootstrapping language-image pre-training with frozen image encoders and large language models. arXiv preprint arXiv:2301.12597  (2023)

\bibitem{Li2022BLIPBL}
Li, J., Li, D., Xiong, C., Hoi, S.: Blip: Bootstrapping language-image pre-training for unified vision-language understanding and generation. In: International Conference on Machine Learning. pp. 12888--12900. PMLR (2022)

\bibitem{Li2023EvaluatingOH}
Li, Y., Du, Y., Zhou, K., Wang, J., Zhao, W.X., Wen, J.R.: Evaluating object hallucination in large vision-language models. arXiv preprint arXiv:2305.10355  (2023)

\bibitem{li2023monkey}
Li, Z., Yang, B., Liu, Q., Ma, Z., Zhang, S., Yang, J., Sun, Y., Liu, Y., Bai, X.: Monkey: Image resolution and text label are important things for large multi-modal models. arXiv preprint arXiv:2311.06607  (2023)

\bibitem{lin2017feature}
Lin, T.Y., Doll{\'a}r, P., Girshick, R., He, K., Hariharan, B., Belongie, S.: Feature pyramid networks for object detection. In: Proceedings of the IEEE conference on computer vision and pattern recognition. pp. 2117--2125 (2017)

\bibitem{cocodataset}
Lin, T.Y., Maire, M., Belongie, S., Hays, J., Perona, P., Ramanan, D., Doll{\'a}r, P., Zitnick, C.L.: Microsoft coco: Common objects in context. In: Computer Vision--ECCV 2014: 13th European Conference, Zurich, Switzerland, September 6-12, 2014, Proceedings, Part V 13. pp. 740--755. Springer (2014)

\bibitem{Liu2023ImprovedBW}
Liu, H., Li, C., Li, Y., Lee, Y.J.: Improved baselines with visual instruction tuning. ArXiv:2310.03744  (2023)

\bibitem{llava}
Liu, H., Li, C., Wu, Q., Lee, Y.J.: Visual instruction tuning. arXiv preprint arXiv:2304.08485  (2023)

\bibitem{liu2023mmbench}
Liu, Y., Duan, H., Zhang, Y., Li, B., Zhang, S., Zhao, W., Yuan, Y., Wang, J., He, C., Liu, Z., et~al.: Mmbench: Is your multi-modal model an all-around player? arXiv preprint arXiv:2307.06281  (2023)

\bibitem{scienceqa}
Lu, P., Mishra, S., Xia, T., Qiu, L., Chang, K.W., Zhu, S.C., Tafjord, O., Clark, P., Kalyan, A.: Learn to explain: Multimodal reasoning via thought chains for science question answering. In: The 36th Conference on Neural Information Processing Systems (NeurIPS) (2022)

\bibitem{marino2019okvqa}
Marino, K., Rastegari, M., Farhadi, A., Mottaghi, R.: Ok-vqa: A visual question answering benchmark requiring external knowledge. In: Proceedings of the IEEE/cvf conference on computer vision and pattern recognition. pp. 3195--3204 (2019)

\bibitem{OpenAI2023ChatGPT}
OpenAI: Chatgpt. \url{https://chat.openai.com} (2023)

\bibitem{OpenAI2023GPT4TR}
OpenAI: Gpt-4 technical report. ArXiv:abs/2303.08774  (2023)

\bibitem{Ouyang2022TrainingLM}
Ouyang, L., Wu, J., Jiang, X., Almeida, D., Wainwright, C., Mishkin, P., Zhang, C., Agarwal, S., Slama, K., Ray, A., et~al.: Training language models to follow instructions with human feedback. Advances in Neural Information Processing Systems  \textbf{35},  27730--27744 (2022)

\bibitem{gpt4llm}
Peng, B., Li, C., He, P., Galley, M., Gao, J.: Instruction tuning with gpt-4. arXiv preprint arXiv:2304.03277  (2023)

\bibitem{peng2023kosmos}
Peng, Z., Wang, W., Dong, L., Hao, Y., Huang, S., Ma, S., Wei, F.: Kosmos-2: Grounding multimodal large language models to the world. arXiv preprint arXiv:2306.14824  (2023)

\bibitem{radford2021learning}
Radford, A., Kim, J.W., Hallacy, C., Ramesh, A., Goh, G., Agarwal, S., Sastry, G., Askell, A., Mishkin, P., Clark, J., et~al.: Learning transferable visual models from natural language supervision. In: International conference on machine learning. pp. 8748--8763. PMLR (2021)

\bibitem{rao2023dynamic}
Rao, Y., Liu, Z., Zhao, W., Zhou, J., Lu, J.: Dynamic spatial sparsification for efficient vision transformers and convolutional neural networks. IEEE Transactions on Pattern Analysis and Machine Intelligence  (2023)

\bibitem{rasley2020deepspeed}
Rasley, J., Rajbhandari, S., Ruwase, O., He, Y.: Deepspeed: System optimizations enable training deep learning models with over 100 billion parameters. In: Proceedings of the 26th ACM SIGKDD International Conference on Knowledge Discovery \& Data Mining. pp. 3505--3506 (2020)

\bibitem{ren2015faster}
Ren, S., He, K., Girshick, R., Sun, J.: Faster r-cnn: Towards real-time object detection with region proposal networks. Advances in neural information processing systems  \textbf{28} (2015)

\bibitem{ronneberger2015u}
Ronneberger, O., Fischer, P., Brox, T.: U-net: Convolutional networks for biomedical image segmentation. In: Medical Image Computing and Computer-Assisted Intervention--MICCAI 2015: 18th International Conference, Munich, Germany, October 5-9, 2015, Proceedings, Part III 18. pp. 234--241. Springer (2015)

\bibitem{textvqa}
Singh, A., Natarajan, V., Shah, M., Jiang, Y., Chen, X., Batra, D., Parikh, D., Rohrbach, M.: Towards vqa models that can read. In: Proceedings of the IEEE/CVF conference on computer vision and pattern recognition. pp. 8317--8326 (2019)

\bibitem{touvron2023llama}
Touvron, H., Lavril, T., Izacard, G., Martinet, X., Lachaux, M.A., Lacroix, T., Rozi{\`e}re, B., Goyal, N., Hambro, E., Azhar, F., Rodriguez, A., Joulin, A., Grave, E., Lample, G.: Llama: Open and efficient foundation language models. arXiv preprint arXiv:2302.13971  (2023)

\bibitem{Touvron2023Llama2O}
Touvron, H., Martin, L., Stone, K., Albert, P., Almahairi, A., Babaei, Y., Bashlykov, N., Batra, S., Bhargava, P., Bhosale, S., et~al.: Llama 2: Open foundation and fine-tuned chat models. arXiv preprint arXiv:2307.09288  (2023)

\bibitem{wang2020deep}
Wang, J., Sun, K., Cheng, T., Jiang, B., Deng, C., Zhao, Y., Liu, D., Mu, Y., Tan, M., Wang, X., et~al.: Deep high-resolution representation learning for visual recognition. IEEE transactions on pattern analysis and machine intelligence (43(10)),  3349--3364 (2020)

\bibitem{wang2021adaptive}
Wang, Y., Chen, Z., Jiang, H., Song, S., Han, Y., Huang, G.: Adaptive focus for efficient video recognition. In: Proceedings of the IEEE/CVF International Conference on Computer Vision. pp. 16249--16258 (2021)

\bibitem{wang2020glance}
Wang, Y., Lv, K., Huang, R., Song, S., Yang, L., Huang, G.: Glance and focus: a dynamic approach to reducing spatial redundancy in image classification. Advances in Neural Information Processing Systems pp. 33, 2432--2444 (2020)

\bibitem{wei2022chain}
Wei, J., Wang, X., Schuurmans, D., Bosma, M., Xia, F., Chi, E., Le, Q.V., Zhou, D., et~al.: Chain-of-thought prompting elicits reasoning in large language models. Advances in Neural Information Processing Systems  \textbf{35},  24824--24837 (2022)

\bibitem{ye2023mplug}
Ye, Q., Xu, H., Xu, G., Ye, J., Yan, M., Zhou, Y., Wang, J., Hu, A., Shi, P., Shi, Y., et~al.: mplug-owl: Modularization empowers large language models with multimodality. arXiv preprint arXiv:2304.14178  (2023)

\bibitem{Yu2023MMVetEL}
Yu, W., Yang, Z., Li, L., Wang, J., Lin, K., Liu, Z., Wang, X., Wang, L.: Mm-vet: Evaluating large multimodal models for integrated capabilities. ArXiv:2308.02490  (2023)

\bibitem{zhu2021dynamic}
Zhu, M., Han, K., Wu, E., Zhang, Q., Nie, Y., Lan, Z., Wang, Y.: Dynamic resolution network. Advances in Neural Information Processing Systems pp. 34, 27319--27330 (2021)

\bibitem{zhu2020deformable}
Zhu, X., Su, W., Lu, L., Li, B., Wang, X., Dai, J.: Deformable detr: Deformable transformers for end-to-end object detection. arXiv preprint arXiv:2010.04159  (2020)

\end{thebibliography}
\end{document}